\documentclass{ifm-tr}

\usepackage{amsmath,amsfonts,bm}

\def\eqref#1{equation~\ref{#1}}

\def\1{\bm{1}}

\DeclareMathAlphabet{\mathsfit}{\encodingdefault}{\sfdefault}{m}{sl}
\SetMathAlphabet{\mathsfit}{bold}{\encodingdefault}{\sfdefault}{bx}{n}

\usepackage{amsmath,amsfonts,amssymb,mathtools,bm}  
\usepackage{thmtools}          
\usepackage{algorithm}         
\usepackage{algpseudocode}     
\usepackage{enumitem}          
\usepackage{siunitx}           
\usepackage{tikz}              
\usepackage{booktabs}          
\usepackage{subcaption}        

\usepackage{hyperref}       
\usepackage{url}            

\usepackage{xurl}          
\usepackage{booktabs}       
\usepackage{amsfonts}       
\usepackage{nicefrac}       
\usepackage{microtype}      
\usepackage{graphicx}
\usepackage{subcaption} 
\usepackage{xspace}
\usepackage[frozencache,cachedir=.]{minted}
\usepackage{longtable}

\usepackage{nicematrix}
\usepackage{multirow}
\usepackage{makecell}
\usepackage{soul}  
\usepackage{wrapfig}
\usepackage{enumitem}

\usepackage{lipsum}

\usepackage{colortbl}
\usepackage[normalem]{ulem}
\usepackage{floatflt}
\usepackage{xspace}

\definecolor{cellHighlight}{HTML}{dbefff}

\newcommand\rurl[1]{%
    \href{https://#1}{\nolinkurl{#1}}%
}

\newcommand{\benchmark}{\texttt{WR-Arena}\xspace}

\newcommand{\eg}{\emph{e.g.,}\xspace}

\newcommand{\ignore}[1]{}

\title{World Reasoning Arena}

\author[]{\textbf{PAN Team, Institute of Foundation Models}}\authorsep{}

\affiliation[]{Mohamed bin Zayed University of Artificial Intelligence}

\reportnumber{2026-03-30}

\abstract{
World models (WMs) are intended to serve as internal simulators of the real world that enable agents to understand, anticipate, and act upon complex environments. Existing WM benchmarks remain narrowly focused on next-state prediction and visual fidelity, overlooking the richer simulation capabilities required for intelligent behavior. To address this gap, we introduce \benchmark, a comprehensive benchmark for evaluating WMs along three fundamental dimensions of next world simulation: (i) Action Simulation Fidelity, the ability to interpret and follow semantically meaningful, multi-step instructions and generate diverse counterfactual rollouts; (ii) Long-horizon Forecast, the ability to sustain accurate, coherent, and physically plausible simulations across extended interactions; and (iii) Simulative Reasoning and Planning, the ability to support goal-directed reasoning by simulating, comparing, and selecting among alternative futures in both structured and open-ended environments. We build a task taxonomy and curate diverse datasets designed to probe these capabilities, moving beyond single-turn and perceptual evaluations. Through extensive experiments with state-of-the-art WMs, 
our results expose a substantial gap between current models and human-level hypothetical reasoning, and establish \benchmark as both a diagnostic tool and a guideline for advancing next-generation world models capable of robust understanding, forecasting, and purposeful action.
The code is available at \url{https://github.com/MBZUAI-IFM/WR-Arena}.
}

\begin{document}
\maketitle



\vspace{0.75cm}

\hfill

\section{Introduction}

A world model (WM) is the algorithmic surrogate of the real-world environment that intelligent agents experience and act upon~\citep{wm-2018,xing2025critiquesworldmodels}. Rather than merely predicting observations, a WM functions as an internal hypothetical simulator capable of representing the manifold possibilities that arise from interactions between an agent and its environment. In this view, a WM supports next world state prediction for \textbf{Next World Simulation}, the ability to generate and evaluate the outcomes of actions under diverse conditions (like visionary simulations in science fiction \emph{Dune}). By mentally exploring alternative futures, a world model enables machines to perform thought experiments that ground reasoning, planning, and decision making. It not only deepens a real-world agent’s understanding of its environment but also provides a foundation for extrapolating knowledge acquired in familiar contexts to novel tasks and complex, previously unseen scenarios.

\begin{figure}[t]
     \centering
     \includegraphics[width=1\linewidth]{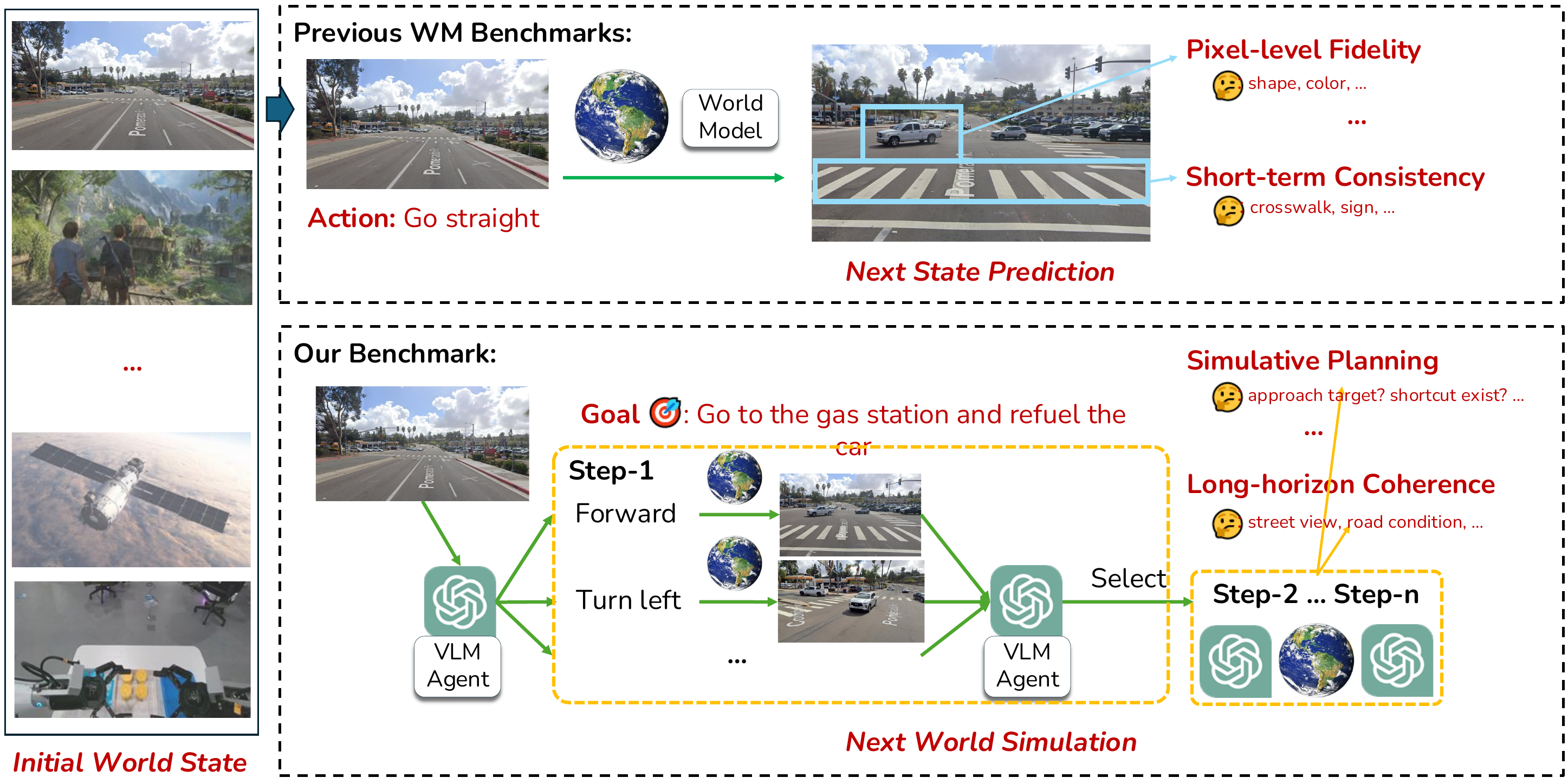}
     \caption{The different focus of existing world model and our evaluation benchmarks.}
     \label{fig:intro}
\end{figure}

In recent years, a number of benchmarks have been proposed to evaluate WMs~\citep{worldmodelbench,duan2025worldscore,gao2025visionlanguagemodelsinternalworld,pbench2024}. 
However, the majority of existing evaluations remain centered on low-level metrics or short-term action/physics simulation (\eg pixel-level fidelity and action reconstruction quality). 
Prior analyses~\citep{xing2025critiquesworldmodels} have noted that many current WMs produce visually plausible outputs yet still fail to respect fundamental physical consistency or long-term scene structure, revealing limitations in their underlying world understanding.
A successful world model should \textbf{maintain a coherent environment} in which objects, agents, and causal dynamics evolve consistently across time, for supporting intelligent behaviors, including high-level action simulation, long-horizon dynamics coherency, and simulative reasoning and planning, as shown in Figure~\ref{fig:intro}.
Without testing these abilities, current benchmarks cannot determine whether a model truly functions as a reliable world simulation sandbox for long-horizon reasoning, decision making, and purposeful action.


To fill this gap, we propose a new evaluation standard that emphasizes three advanced capabilities of WMs: (1) \emph{Action Simulation Fidelity:} follow semantically meaningful, high-level instructions targeting agents or environments, and generate diverse rollouts (\eg video clips). This evaluates whether a WM can simulate abstract commands into coherent observation trajectories. (2) \emph{Long-horizon Forecast:} sustain accurate and reasonable rollouts across extended sequences, minimizing error accumulation and preserving coherence over time. This tests the stability of WMs in multi-round interactions. (3) \emph{Simulative Reasoning and Planning:} support goal-directed reasoning by simulating different rollouts for comparing alternative futures in both structured and open-ended environments. This measures whether WMs can act as active planners rather than passive predictors, \eg simulating several possible routes before the agent choosing the best.

\begin{figure}[!t]
     \centering
     \includegraphics[width=1\linewidth]{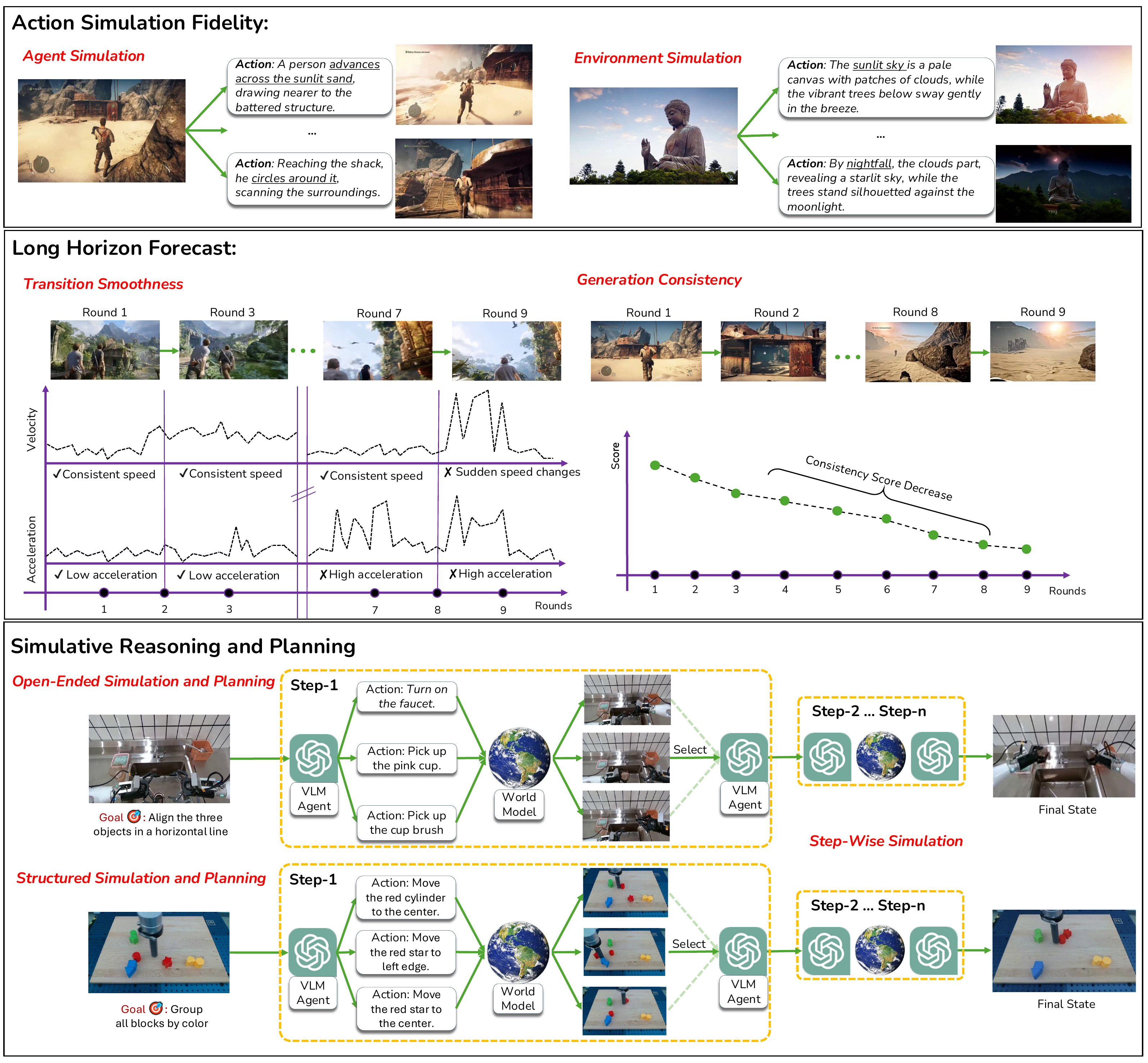}
    \caption{Taxonomy with detailed examples of our evaluation benchmark.}
     \label{fig:taxonomy}
\end{figure}

In this paper, we introduce \benchmark, a comprehensive benchmark that systematically evaluates WMs across these three dimensions. Building on them, we construct a taxonomy of evaluation tasks (Figure~\ref{fig:taxonomy}) that captures both fine-grained skills and their intersections across diverse environments. For each dimension, we curate test datasets and design evaluation protocols that extend beyond perceptual fidelity to test reasoning, interaction, and planning capabilities.

Using \benchmark, we conduct a large-scale evaluation of state-of-the-art WMs, leading to several key findings:

$\bullet$ Current models struggle with action simulation control, especially for environment-centric commands, revealing a significant gap in faithfully following high-level instructions.

$\bullet$ Long-horizon simulation remains difficult for all models, with error accumulation degrading consistency over extended rollouts.

$\bullet$ Only world models that produce semantically actionable rollouts significantly improve planning performance, showing that perceptual quality alone is insufficient for decision making.

$\bullet$ Models that jointly optimize understanding, prediction, and control (\eg PAN) deliver the most balanced performance across evaluation dimensions.



These results reveal a substantial gap between current WMs and human-level hypothetical reasoning. At the same time, they highlight the promise of our benchmark \benchmark as both a diagnostic tool and a guideline for the development of next-generation world models that can understand, forecast, and plan in complex real-world environments.


\section{Preliminaries}\label{sec:prelims}

\subsection{World Model}\label{subsec:wm_def}

\paragraph{Definition.}
A \emph{world model} (WM) is a generative model that simulates possible futures across diverse domains, including the physical, mental, social, and evolutionary worlds. Operationally, a WM takes as input a previous world state $s$ and an action $a$, and produces the next state $s'$ through a transformation function:
\begin{equation}
    s' \sim p(s' \mid s, a). \tag{1}
\end{equation}
By iteratively applying this transition function, a WM can generate trajectories that represent how the world might evolve under different action sequences. This ability enables machines to perform \emph{thought experiments}: internally simulating alternative scenarios, including counterfactual ones, and evaluating which trajectories best achieve a given goal.
This definition parallels the cognitive hypothesis that humans reason not only by applying explicit rules but also by simulating outcomes with internal mental models \citep{johnson2010mental}. For example, rather than acting purely through deterministic optimization, humans often project multiple possible futures (\eg imagining whether helping someone in distress leads to gratitude, self-exhaustion, or no change ) and then act based on the expected reward of those futures \citep{xing2025critiquesworldmodels}. WMs aim to endow machines with this same capability to “see the future,” supporting more flexible and adaptive reasoning.


\paragraph{World Model for Simulative Reasoning and Planning.}
The most fundamental training paradigm for WMs is \emph{next state prediction}, in which the model learns to minimize the difference between predicted and observed states.
Beyond next state prediction, WMs enable \emph{next world simulation} for reasoning and decision-making by generating and comparing alternative futures. Given an initial state $s$ and a goal $g$, an agent can generate candidate action sequences $\langle a_1,\dots,a_T\rangle$, roll them out through the WM, and select the trajectory $s_{1:T}$ that best achieves $g$. This approach supports both long-horizon planning (\eg real-world robotic control) and high-level reasoning (\eg evaluating counterfactual scenarios in open-world environments).
Crucially, WMs also facilitate transfer and generalization across domains, since many real-world dynamics share the underlying mechanistic regularity. Just as humans use prior embodied experience to adapt to novel situations (\eg a scuba diver adapting to low gravity when walking on the moon), machines can use WMs to extend past knowledge to unfamiliar tasks. This makes WMs not only predictive simulators but also foundations for zero-shot adaptation, robust decision-making, and complex planning in unstructured environments.

\subsection{World Model Evaluation}\label{subsec:wm_eval}

\paragraph{Limitations of Existing Benchmarks.}
Most existing world model benchmarks emphasize short-term state prediction or visual fidelity rather than deeper reasoning and control \citep{worldmodelbench, pbench2024}. They typically measure three aspects:
(1) \emph{Low-level control}: such as matching immediate motor commands to resulting motions;
(2) \emph{Next-state prediction}: whether the model correctly predicts the very next frame or state given current input; and
(3) \emph{Video fidelity}: which captures perceptual realism of generated outputs.
While these criteria are useful, they largely assume a single-turn setting and neglect long-horizon or multi-round interactions. As a result, they fail to capture key demands in real-world applications like embodied AI, or autonomous driving, where models must sustain coherent trajectories, respect physical constraints, and follow instructions over extended time horizons. Similarly, video generation benchmarks often overemphasize perceptual capability without testing whether models consistently track agents, enforce causal dynamics, or maintain instruction alignment throughout rollouts.


\paragraph{Our Focus.}
To address these gaps, our benchmark is designed to evaluate \emph{advanced capabilities} of a competent world model. Rather than stopping at short-term prediction or visual quality, we assess whether models can simulate, reason, and act in realistic long-horizon scenarios. Concretely, our evaluation framework consists of three complementary dimensions:


\begin{itemize}
\item \textbf{Action Simulation Fidelity:} Tests whether a model can execute semantically meaningful, multi-step instructions that target either agents or environments, producing diverse and counterfactual futures from the same starting state.
\item \textbf{Long-horizon Forecast:} Measures the ability to sustain coherent rollouts over extended action sequences, evaluating prediction accuracy, temporal smoothness, and error accumulation beyond short horizons.
\item \textbf{Simulative Reasoning and Planning:} Assesses whether models can support goal-directed reasoning and planning, both in structured (local) and unstructured (open-world) environments, through iterative simulation of candidate actions as ``thought experiments''.
\end{itemize}

By unifying these three aspects, our suite moves beyond single-turn or purely perceptual testing and provides a holistic picture of how a world model \emph{maps} the current state, \emph{rolls} it into possible futures, and \emph{acts} to achieve goals in complex, realistic environments.


\section{Evaluation Framework}
We present a comprehensive framework for systematically evaluating world models, covering the fundamental and advanced capabilities.
In this work, we mainly focus on testing the advanced capabilities that are essential for real-world application, namely Action Simulation Fidelity, Long-horizon Forecast, and Simulative Reasoning and Planning.
These three dimensions assess whether a model can serve as a reliable simulator of the real world, to handle the complex goal-directed application like auto-driving and robot navigation.

\subsection{Action Simulation Fidelity}

Action Simulation Fidelity is the property of a world model to accurately follow semantically specified, multi-step natural language instructions, \eg cook a dish and drive back home.
Here, given an initial state and high-level control instructions, we evaluate if the model could generate a sequence of reasonable states that faithfully follow the control to accomplish this task.
Concretely, given an initial world state $s_0$, we employ the LLM (e.g., GPT-4o \citep{openai2024gpt4}) to propose several multi-step high-level action sequences $\mathcal{A}=\langle a_1,\cdots,a_n\rangle$ under simple feasibility constraints (\eg non-contradictory and causally applicable). Then, the world model simulates a rollout $\mathcal{R}(s_0,\mathcal{A})=\langle s_1,\dots,s_T\rangle$ conditioned on $\mathcal{A}$. 
We score these simulations using vision-language models as judge following existing protocols~\citep{worldmodelbench}, focusing on action faithfulness and action precision.
Based on the above design, we instantiate two settings that differ only in \emph{who} or \emph{what} the controls target: the agent versus the environment.

\paragraph{Agent Simulation.}
Agent Simulation evaluates whether the model can drive the \emph{controllable entity} through intended high-level behaviors while keeping the background dynamics stable. 
For each $s_0$ we sample multiple distinct $\mathcal{A}$ to induce counterfactual futures.
The assessment verifies how the world model maintains faithful simulation of agent control actions while producing appropriately diverse outcomes across different sequences. This capability to generate multiple coherent futures from a single starting state is essential for real-world planning applications with different action strategies comparison.

\paragraph{Environment Simulation.}
Environment Simulation evaluates whether the model can apply high-level \emph{scene} interventions and simulate their causal consequences, while the agent’s policy remains neutral (\eg continue forward). Each sequence contains scene-level actions that are visually verifiable and have predictable downstream effects.
We follow the evaluation setting of agent control and also consider both the accurate simulation and multi-future diversity.

\subsection{Long-horizon Forecast}
Long-horizon forecast refers to the property of a world model to maintain coherent, high-quality simulations over extended sequences of interactions. Beyond short-term accuracy, this dimension evaluates whether models can avoid error accumulation and degradation when reasoning many steps about the future. Concretely, given an initial world state $s_0$ and a sequence of actions $\mathcal{A}=\langle a_1,\cdots,a_n\rangle$ spanning multiple rounds, the model simulates a rollout $\mathcal{R}(s_0,\mathcal{A})=\langle s_1,\dots,s_T\rangle$ that should remain visually faithful, dynamically plausible, and consistent throughout the horizon. 
Thus, we test it from the following two dimensions, namely transition smoothness, and generation consistency.

\paragraph{Transition Smoothness.}
To probe multi-step dynamics, we extend each initial state with a sequence of action spanning $k$ rounds. The model is expected to produce smooth and physically plausible trajectories without sudden jumps or artifacts as the sequence unfolds. We quantify transition smoothness using optical flow continuity across frames, measuring whether trajectories evolve consistently over time. This evaluation emphasizes whether models can sustain coherent dynamics when actions extend beyond the short horizon, avoiding sudden transitions or implausible discontinuities.

In specific, we score the \emph{transitions} at each round boundary. 
Around each round boundary we take a short symmetric window and summarize framewise motion with two signals: (i) a velocity proxy ($v_t$) from optical flow to ensure there is perceptible motion, and (ii) its finite-difference acceleration ($a_t$) to penalize abrupt changes. 
The metric returns high scores only when motion is present \emph{and} changes are gradual: static segments (low $v_t$) and abrupt cuts (high $a_t$) should both score low. 
We then form a per-boundary score that \emph{rewards} visible motion but \emph{exponentially down-weights} jerks; per-video normalization makes scores comparable across scenes. 
Averaging these boundary scores yields a single \emph{Multi-round Smoothness} (MRS) score, higher when actions are simulated smoothly and lower when transitions are static or twitchy. 

Formally, let the rollout frames be $\langle I_1,\dots,I_T\rangle$ split into $k$ rounds with boundaries $\{b_1,\dots,b_{k-1}\}$. 
Using dense optical flow between $I_t$ and $I_{t+1}$, denote per-pixel flow $\mathbf{u}_t(p)$ and magnitude $\|\mathbf{u}_t(p)\|_2$. 
Define framewise velocity and acceleration as

\[
v_t=\|\mathbf{u}_t(p)\|_2,\qquad
a_t=\| v_t-v_{t-1}\|_2\;(t\ge2).
\]
For each boundary $b_r$, take a symmetric window $\mathcal{B}_r$ covering the last and first $\delta$ fraction of frames of rounds $r$ and $r{+}1$ (we use $\delta=0.10$). 
Let $\tilde v_t$ and $\tilde a_t$ be per-video normalized versions of $v_t$ and $a_t$ obtained by dividing by the 99th percentile and clipping to $[0,1]$. 
The per-boundary smoothness is
\[
S_r=\frac{1}{|\mathcal{B}_r|}\sum_{t\in\mathcal{B}_r}
\tilde v_t\,\exp\!\ \bigl(-\lambda\,\tilde a_t\bigr),
\quad\text{with }\lambda=2.5,
\]
and the overall \emph{Transition Smoothness} is the average over boundaries,
\[
\mathrm{MRS}=\frac{1}{k-1}\sum_{r=1}^{k-1}S_r\qquad(\uparrow\text{ is better}).
\]
This construction isolates where discontinuities most often occur (round hand-offs), avoids the trivial “smoothness” of being still (via $\tilde v_t$), and aligns with perceptual sensitivity to jerks (via the exponential penalty), while remaining comparable across diverse videos (via normalization).

\paragraph{Generation Consistency.}
Finally, we assess the cumulative robustness of long-horizon rollouts. Starting from collected world states across diverse domains, we apply $k$-round action sequences and measure quality across the entire rollout. Following WorldScore~\citep{duan2025worldscore}, we track two key metrics: content alignment and style consistency. To highlight error accumulation, we apply additive penalties that amplify small inaccuracies as rounds progress, yielding a weighted average score that reflects both early-step accuracy and long-term stability. A strong world model should minimize compounding errors, preserving visual fidelity and physical coherence even after extended interactions.

Formally, if the normalized scores ($0-100$) for content alignment and style consistency are aggregated for $k$ rounds, and listed as $\{s_1,\dots,s_{k}\}$, then the additive penalty that measures the decrease in the performance is given by:
\[
\mathrm{AP}_\lambda(s_1{:}s_k) = s_1 \exp \left( -\lambda \frac{1}{k}\sum_{t=1}^{k} \lvert s_t - s_1\rvert \right)
\]
When there is no degradation, $\mathrm{AP}_\lambda$ is close to $s_1$. As the average deviation from the start grows, the score falls off exponentially, tying \emph{initial fidelity} to \emph{rollout stability}.

\subsection{Simulative Reasoning and Planning}
Simulative Reasoning and Planning highlights the role of world models as active simulators for goal-directed behavior, which should not only predict faithful outcomes of actions but also support iterative decision-making in complex planning and reasoning contexts.
Specifically, given an initial world state $s_0$ and a goal $g$, the model must generate a rollout $\mathcal{R}(s_0,g)=\langle s_1,\dots,s_T\rangle$ where each intermediate state reflects progress toward the goal. 
We evaluate whether the world model can collaborate with a vision-language model (VLM) to plan in natural language space. For the latter, the VLM acts as the planner to iteratively propose candidate actions, and the world model simulates their outcomes, where the planner (i.e., VLM) will consider world models' simulations and select the best corresponding action to advance toward the goal $g$.


\paragraph{Step-Wise Simulation.}
Step-Wise Simulation evaluates the world model's predictive capability on the immediate consequence from a given action, and serves as the foundation of long-horizon simulative reasoning and planning. At each step of a long-horizon prediction task, the model should simulate the next world observation that faithfully reflects the commanded action and all resulting consequences. We evaluate this capability using robot arm manipulation tasks from WM-ABench~\citep{gao2025visionlanguagemodelsinternalworld}, where each instance presents an initial observation and action, and the model needs to select the correct next observation from one ground-truth target and three carefully curated distractors. 
For generative models that produce next observations such as PAN, we employ human assessments on the predicted observations, examining whether object relationships and physical effects align with the ground-truth next observation. For embedding-based models such as V-JEPA 2~\citep{assran2025vjepa2selfsupervisedvideo}, we compute the similarity between the predicted latent world state and the latent state from the ground-truth observation. To ensure domain alignment, all models are finetuned on a similar robotic manipulation dataset (i.e., Agibot ~\citep{bu2025agibot}). To enable V-JEPA 2 processing language-based actions, we extend it with the UMT5 encoder~\citep{umt5} from WAN2.1~\citep{wan2025wan} and then conduct the finetuning. This task captures the foundational causal reasoning capability underlying the multi-step simulative planning.

\paragraph{Open-Ended Simulation and Planning.}
Open-Ended Simulation and Planning evaluates whether world models can reason and act in complex, naturalistic environments. In this task, robots operate in realistic household settings and must interact with everyday objects under diverse and unpredictable contexts. Successful planning requires multi-step reasoning and long-horizon foresight. In each iteration, a VLM agent (e.g., OpenAI-o3 ~\citep{OpenAI2025o3o4minisystemcard}) will first propose candidate actions from the current observation, then the world model will simulate the consequences of these actions, and finally the VLM agent will select the action that has the predicted next observation closest to the goal. This iterative process continues until the task goal is achieved or the planning budget is reached. We curate 15 scenarios from the Agibot~\citep{bu2025agibot} dataset for evaluation, with human assessment on trajectory-level completion and simulation quality. For V-JEPA 2, we conduct image editing on the initial observations to create plausible goal observations. This task evaluates whether models can generalize planning ability to diverse, open-ended environments.

\paragraph{Structured Simulation and Planning.}
Structured Simulation and Planning focuses on controlled, structured settings where complexity is reduced but precise reasoning is still required. We use the tabletop setting where robots manipulate regular objects such as colored cubes and spheres from the Language Table dataset~\citep{lynch2022interactivelanguagetalkingrobots}. This structured environment minimizes confounding variability, enabling a focused study of language-grounded reasoning and fine-grained manipulation. Following the same agent–world model iterative planning loop, we curate 47 cases from selected observations in the Language Table dataset~\citep{lynch2022interactivelanguagetalkingrobots}. Our task cases cover different types of spatial arrangements such as \textit{grouping the blue objects} and \textit{aligning the objects into a horizontal line}. Again we conduct blinded human assessments on both goal achievement and trajectory quality. This task complements open-ended simulation and planning by providing a simplified testbed that focusing on a model’s reasoning and manipulation capabilities under well-defined conditions. 
\section{Experiment}
\label{sec:experiment}

\subsection{Baselines}
We conduct comprehensive evaluation on world models designed for interactive simulation and reasoning, and video generation models (both open-source and closed-source commercial APIs).

\paragraph{World Models.}
World models are designed to predict future world observations conditioned on actions, enabling agents to reason about consequences and plan accordingly.

$\bullet$ \textbf{Cosmos (NVIDIA)}~\citep{nvidia2025cosmosworldfoundationmodel}: a family of ``world foundation models'' aimed at training robots and autonomous systems via photo-realistic video and synthetic data generation. It is positioned for world dynamics and control rather than purely creative video synthesis.

$\bullet$ \textbf{V-JEPA2}~\citep{assran2025vjepa2selfsupervisedvideo}: Meta's video JEPA line models latent video dynamics via masked prediction in latent space, with strong motion understanding and forecasting.

$\bullet$ \textbf{PAN}~\citep{panteam2025panworldmodelgeneral}: a general, interactable, and long-horizon world model that predicts future world states through high-quality video simulation conditioned on history and natural language actions. PAN employs a Generative Latent Prediction (GLP) architecture that combines an autoregressive latent dynamics backbone based on a large language model with a video diffusion decoder, enabling open-domain, action-conditioned simulation with coherent long-term dynamics.

\paragraph{Video Generation Models.}
Video generation models focus on producing high-quality visual sequences, typically operating in a prompt-to-video manner. For open-source models, we select WAN-2.1 and WAN-2.2, and for closed-source models, we use KLING, MiniMax, and Gen-3.

$\bullet$ \textbf{WAN-2.1 \& 2.2}~\citep{wan2025wanopenadvancedlargescale}: diffusion-Transformer video generators released by the WAN team. The 2.1 and 2.2 updates emphasize quality and long-range temporal coherence.

$\bullet$ \textbf{KLING}~\citep{klingai-app-cn}: a text/image/video-to-video model capable of 1080p, up to 2-minute clips, designed for cinematic camera controls and strong photorealism. It is broadly adopted in creative workflows.

$\bullet$ \textbf{MiniMax}~\citep{hailuo-ai-minimax}: MiniMax's production video-generation agent platform (text/image conditioned) integrated into its broader agent ecosystem. Its public materials emphasize end-to-end agentic creation rather than low-level model specs.

$\bullet$ \textbf{Gen-3}~\citep{runway-gen3-alpha}: Runway's latest production model with improved motion fidelity, expressive characters, and stronger camera control, positioned for professional media pipelines.

\begin{table}[h]
\centering
\newcommand{\NA}{\cellcolor[HTML]{EFEFEF}}
\small
\renewcommand{\arraystretch}{1.15}
\setlength{\tabcolsep}{4pt}
\begin{tabular}{lcccccccc}
\toprule
\rowcolor[HTML]{E6ECF2}
& \multicolumn{2}{c}{\textbf{Action Simulation Fidelity}} 
& \multicolumn{2}{c}{\textbf{Long-horizon Forecast}}
& \multicolumn{4}{c}{\textbf{Simulative Reasoning \& Planning}} \\[2pt]
\rowcolor[HTML]{E6ECF2}
\textbf{Model} &
\makecell{Agent\\Simulation} &
\makecell{Env.\\Simulation} &
\makecell{Transition\\Smoothness} &
\makecell{Simulation\\Consistency} &
\makecell{Step-Wise \\ Sim.} &
\makecell{Open-Ended\\Sim.} &
\makecell{Struct.\\Sim.} \\
\cmidrule(lr){2-3}\cmidrule(lr){4-5}\cmidrule(lr){6-8}
		
KLING                     & 51.0      & 43.3  & 40.5 & \underline{59.2}     & \NA    & \NA      & \NA \\
MiniMax                   & \textbf{72.3}      & \textbf{51.7}  & 30.7 & 57.2     & \NA    & \NA      & \NA \\
Gen-3                     & 45.3      & \underline{47.3}  & \underline{47.8} & 45.1     & \NA    & \NA      & \NA \\
Cosmos1-14B               & 37.7      & 37.7  & 31.6 & 54.0     & 7.8   & \underline{+6.7\%}   & -2.1\% \\
Cosmos2-14B               & 58.0      & 44.0  & 17.3 & \underline{59.2}     & \underline{31.1}   & +0\%     & +4.3\%  \\
V-JEPA2                   & \NA       & \NA   & \NA  & \NA      & 15.5   & -6.7\%   & \underline{+10.6\%} \\
WAN 2.1                   & 53.7      & 37.0  & 11.1 & 35.8     & \NA    & \NA      & \NA \\
WAN 2.2                   & 65.0      & 39.3  & 20.5 & 37.6     & \NA    & \NA      & \NA \\
PAN                       & \underline{70.3}      & 47.0  & \textbf{53.6} & \textbf{64.1}     & \textbf{56.1}   & \textbf{+26.7\%}  & \textbf{+23.4\%} \\
\bottomrule
\end{tabular}
\vspace{6pt}
\caption{Model performance on Action Simulation Fidelity, Long-horizon Forecast and Simulative Reasoning \& Planning tasks. Results in \textit{Action Simulation Fidelity} are measured based on the criteria proposed in \cite{worldmodelbench} and then normalized. Task success rates in \textit{Open-Ended Sim.} and \textit{Struct. Sim.} are measured in the trajectory level and represent improvements over the pure VLM agent without integrating WM. The best-performing model in each dimension is marked in \textbf{bold}, and the second best is \underline{underlined}.}
\label{tab:long-horizon_plan}
\end{table}

\subsection{Main Results}
Table~\ref{tab:long-horizon_plan} reports results for all the evaluated world models and video generators across our three evaluation dimensions. Overall, maintaining long-horizon transition smoothness and consistency while applying fine-grained environment changes remains difficult for current models.

\paragraph{Action Simulation Fidelity.}
Across all evaluated models, we observe a consistent performance gap between Agent Simulation and Environment Simulation, where performances on agent-centric manipuations exceed those on environment-level manipulations by 11.5\% on average. Notably, no model surpasses 60\% accuracy on environment simulation, suggesting that faithfully modeling scene-level interventions remains a fundamental limitation of current models.
Among all baselines, MiniMax achieves the strongest overall performance (72.33\% on Agent Simulation, 51.67\% on Environment Simulation), though there remain a substantial room for improvement. While prior work~\citep{brooks2024video} suggested that pretrained video generators can function as general-purpose world models, our results show that WAN2.1, trained exclusively on broad video corpora without action-conditioned supervision, exhibits notably weaker simulation fidelity. By contrast, PAN's targeted fine-tuning on action–state aligned sequences yields substantial improvements over WAN2.1, with gains of +16.66\% on Agent Simulation and +10.0\% on Environment Simulation.
These findings underscore two key observations. First, sequence-level action grounding remains a challenging capability for current world models. Second, explicit alignment between action representations and state transitions is essential for high-fidelity simulation, particularly for environment-centric manipulations where all evaluated models exhibit the greatest difficulty.

\begin{figure}[htbp]
  \centering
    \includegraphics[width=0.8\linewidth]{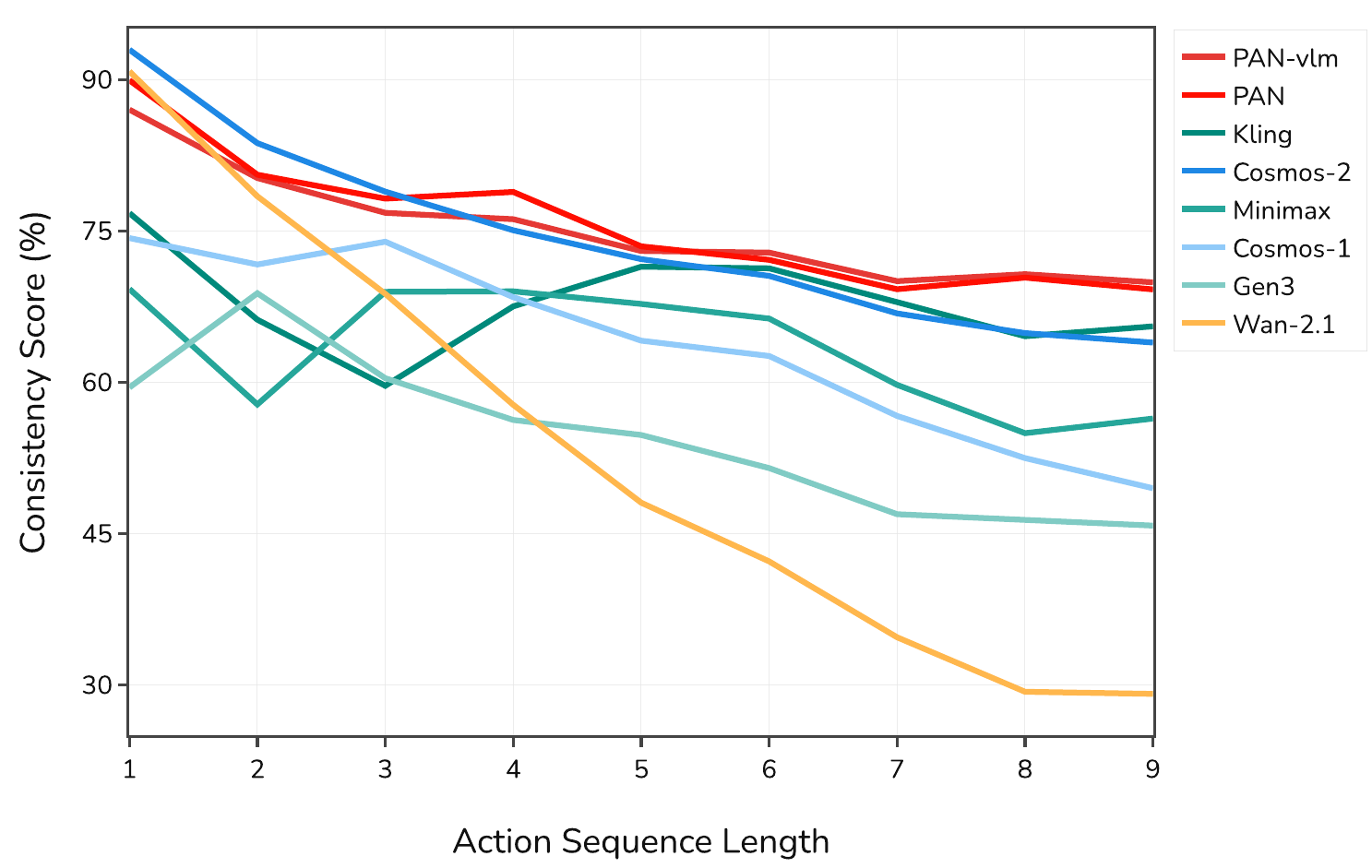}
    \vspace{-5pt}
    \caption{The generation consistency curve \emph{w.r.t.} the action sequence length. Eight world models or video generators execute identical 9-round action sequences from the same initial observations, with consistency scores measured at each round. Most models achieve lower than 75\% consistency after round 5 or 6, indicating fundamental limitations for long-horizon planning applications.
    }
  \label{fig:error_accumulation_rounds}
\end{figure}

\paragraph{Long-horizon Forecast.}
This dimension evaluates whether models can predict future observations over many turns without accumulating blur, while preserving temporal coherence and motion smoothness. Across all evaluated systems, no model exceeds 65\% on either Transition Smoothness or Generation Consistency, underscoring that error accumulation remains a fundamental challenge for long-horizon simulation. As illustrated in Figure~\ref{fig:error_accumulation_rounds}, even the strongest systems exhibit measurable per-round degradation during multi-turn generation.
Among all baselines, PAN achieves the best performance on both metrics, maintaining coherent dynamics across extended sequences. These results suggest that PAN's long-context conditioning provides effective regularization for next-state generation. PAN also clearly surpasses commercial video generation models such as KLING (59.15\%) and MiniMax (57.17\%) on Generation Consistency, maintaining higher content alignment and style stability across turns (Figure~\ref{long-consistency-analysis}). By contrast, WAN2.1 performs poorly on long-horizon metrics, tending to exaggerate motion magnitudes and producing jittery, non-smooth trajectories. As shown in Figure~\ref{long-consistency-analysis}, WAN2.1 exhibits pronounced visual drift as the number of turns increases. We provide a fine-grained, round-by-round breakdown of these results in Section~\ref{long-consistency-analysis}.

\paragraph{Simulative Reasoning and Planning.}
This dimension evaluates whether world models can serve as internal simulators that support goal-directed reasoning and planning, effectively functioning as engines for thought experiments. Specifically, we assess whether models can generate plausible future observations that enable a VLM planner to explore alternative action paths and select actions with foresight. On both \emph{Open-Ended} and \emph{Structured} settings, PAN yields the largest improvements when integrated with the same VLM planner, achieving 26.33\% gain in trajectory-level success over the VLM-only baseline for open-ended settings and 23.40\% in structured environments. In contrast, Cosmos 1 \& 2 and V-JEPA 2 exhibit inconsistent effects that occasionally provide modest improvements but sometimes fail to provide simulations that can serve as guidance for planning. These results suggest that an effective world model must not only produce visually coherent simulations but also generate state transitions that are semantically grounded and causally informative for downstream planners. Among all evaluated models, only PAN demonstrates the potential for reliable counterfactual thought experiments to benefit multi-step planning.

In summary, across all evaluated dimensions, no single baseline dominates. Commercial video generators (KLING, MiniMax, Gen-3) deliver strong perceptual quality and reasonable short-horizon control, yet their closed training and weak domain adaptation limit usefulness for planning on specific dimension. Conversely, embedding-centric or purely generative open-source models (V-JEPA-2, Cosmos) better support internal rollouts for planning but lag in faithful control and temporal stability, leading to brittle downstream execution. In contrast, PAN, grounded by VLM priors and action–state aligned finetuning, strikes the most balanced profile. It sustains good semantic understanding, maintains coherent multi-turn predictions, and follows high-level instructions reliably enough to improve planner success. These findings argue for a unified architectures that co-optimize understanding, prediction, and control as coupled objectives, rather than treating them as separable modules.

In summary, no single baseline achieves dominant performance across all evaluated dimensions. Commercial video generators (KLING, MiniMax, Gen-3) exhibit strong visual fidelity and reasonable capability on short-horizon action simulation; however, their limitation in domain adaptation constrains the utility for task-specific planning. Conversely, open-source embedding-based or generative world models (V-JEPA 2, Cosmos) can be adapted to conduct domain specific simulations but yield inconsistent gains in downstream planning, suggesting that their simulations lack the semantic grounding necessary for effective decision making. PAN, by contrast, achieves the most balanced performance profile across all dimensions. By grounding simulation in VLM priors and employing action–state aligned fine-tuning, PAN maintains robust semantic understanding, coherent multi-turn predictions, and reliable instruction following capabilities, collectively enabling reliable simulations in the planning process. These findings motivate the development of unified architectures that jointly optimize understanding, prediction, and control as tightly coupled objectives, rather than treating them as independent modules.

\subsection{Analysis}
\label{long-consistency-analysis}

\paragraph{Long-horizon Generation Consistency.}

As shown in Table~\ref{tab:long-horizon_plan}, maintaining generation consistency over extended action sequences remains a significant challenge for current world models and video generators, primarily due to error accumulation during multi-step observation prediction. To further investigate this phenomenon, we conduct a fine-grained analysis by measuring per-round consistency across nine consecutive action steps for all models.
We observe that generation consistency degrades monotonically as action sequence length increases across all models, confirming that error accumulation poses a fundamental bottleneck for long-horizon simulation. The severity of degradation, however, varies substantially across models. For instance, WAN 2.1 exhibits the most severe decline, with consistency dropping from approximately 90\% to 30\% over nine rounds. MiniMax, Cosmos-1, and Gen3 similarly fall below 60\% after round 7, indicating limited capacity for sustained simulation in extended planning scenarios such as complex navigation tasks~\citep{xing2025critiquesworldmodels}.
Notably, while PAN does not achieve the highest consistency in the first few rounds, its degradation curve remains flat compared to all other models. After round 3, PAN maintains the highest consistency through the remainder of the action sequence. This slow decay rate suggests that PAN effectively mitigates error accumulation over extended horizons. We attribute this stability to PAN's self-forcing training strategy, which enforces local consistency between neighboring frames and thereby reducing compounding errors over extended simulations.

\section{Qualitative Results}

\begin{figure}[H]
     \centering
     \includegraphics[width=1\linewidth]{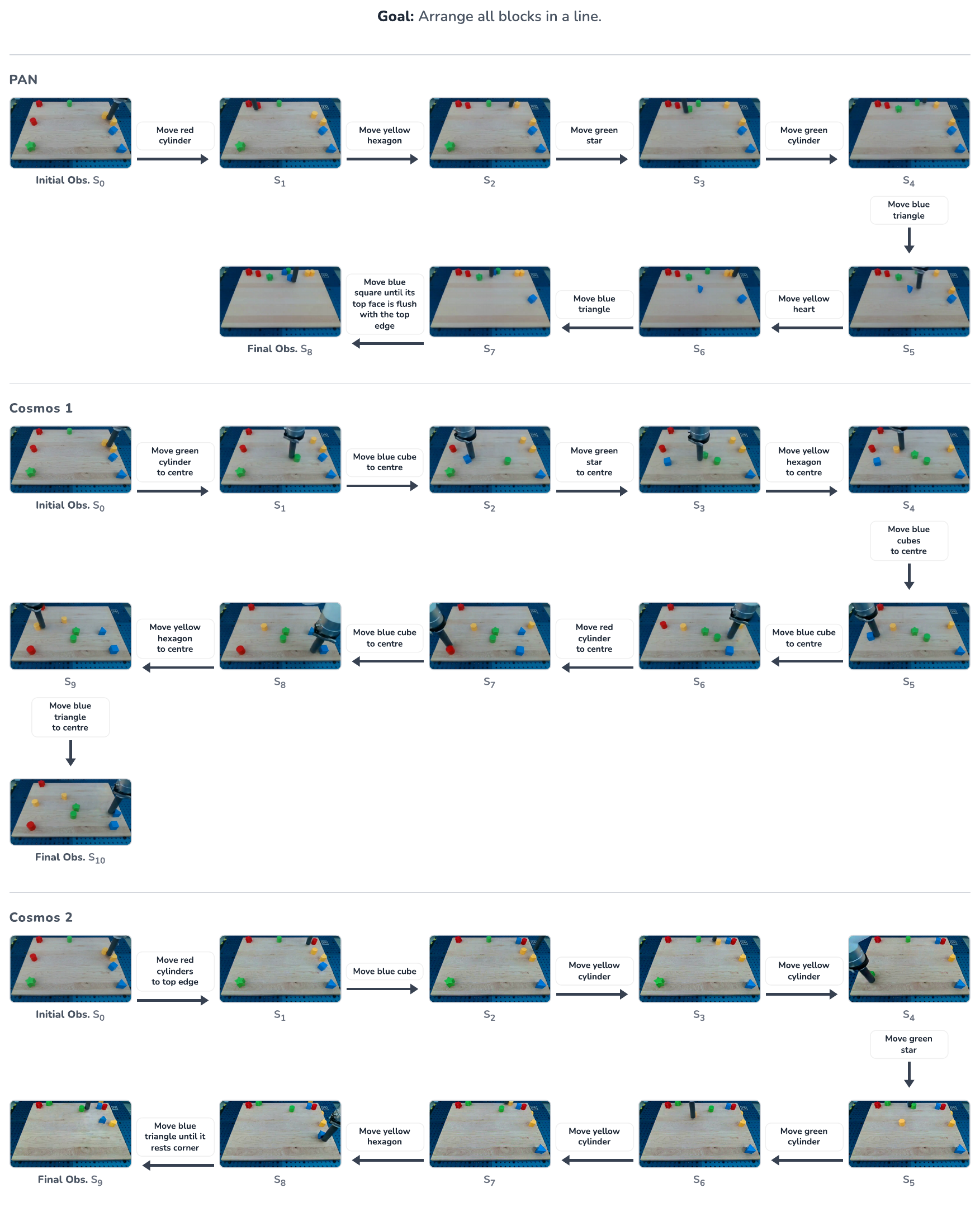}
    \caption{\textbf{Qualitative comparison on Structured Simulation and Planning.} Starting from the same tabletop observation ($S_0$) with the goal of arranging all blocks into a line, PAN, Cosmos 1, and Cosmos 2 each simulate multi-step action plans through an iterative VLM–world model planning loop.}
     \label{fig:case-simulative}
\end{figure}

We present a few examples generated by SOTA world models from our benchmark, and highlight the different focuses of all the evaluated dimensions.

\subsection{Simulative Reasoning and Planing}
Figure~\ref{fig:case-simulative} contrasts how different world models organize multi-step plans under a shared initial observation $S_0$ and a concrete spatial goal (arrange all blocks into a line). Overall, this case study highlights the \emph{structured simulation and planning} dimension: whether a model can (i) decompose the goal into a coherent sequence of atomic actions, (ii) maintain stateful consistency across intermediate predictions, and (iii) converge to a goal-satisfying terminal configuration.

PAN produces a goal-directed and diverse action sequence. Starting from $S_0$, it proposes a series of targeted moves that progressively reduce clutter and align pieces toward the intended line arrangement (\emph{e.g.,} relocating distinct objects such as the red cylinder, yellow hexagon, and green star before resolving remaining misplacements). Notably, PAN’s rollout demonstrates \emph{state-aware refinement}: later steps correct residual issues left by earlier placements (\emph{e.g.,} adjusting a blue square with an explicit geometric constraint such as being flush to an edge), reflecting iterative replanning rather than a one-shot script.
By comparison, Cosmos~1 exhibits a more repetitive planning pattern. Its actions frequently collapse into a single heuristic (moving different objects “to centre”), yielding a longer horizon with weaker evidence of goal decomposition. While such centering behaviors may simplify dynamics and reduce prediction uncertainty, they are less aligned with the explicit goal structure of “forming a line,” and can introduce unnecessary steps that dilute planning efficiency. Cosmos~2 presents an intermediate behavior: it generates more heterogeneous actions than Cosmos~1 (including edge- and corner-related adjustments), suggesting stronger spatial reasoning signals. However, parts of its sequence still appear only loosely coupled to the final arrangement objective, indicating that the model can simulate plausible state transitions but may not consistently prioritize goal progress at each step.

Taken together, this example emphasizes that strong simulative planning is not only about producing visually plausible intermediate states, but also about maintaining a \emph{goal-conditioned action policy} over long horizons—balancing progress, correction, and efficiency under iterative VLM–world model rollouts.

\begin{figure}
     \centering
     \includegraphics[width=1\linewidth]{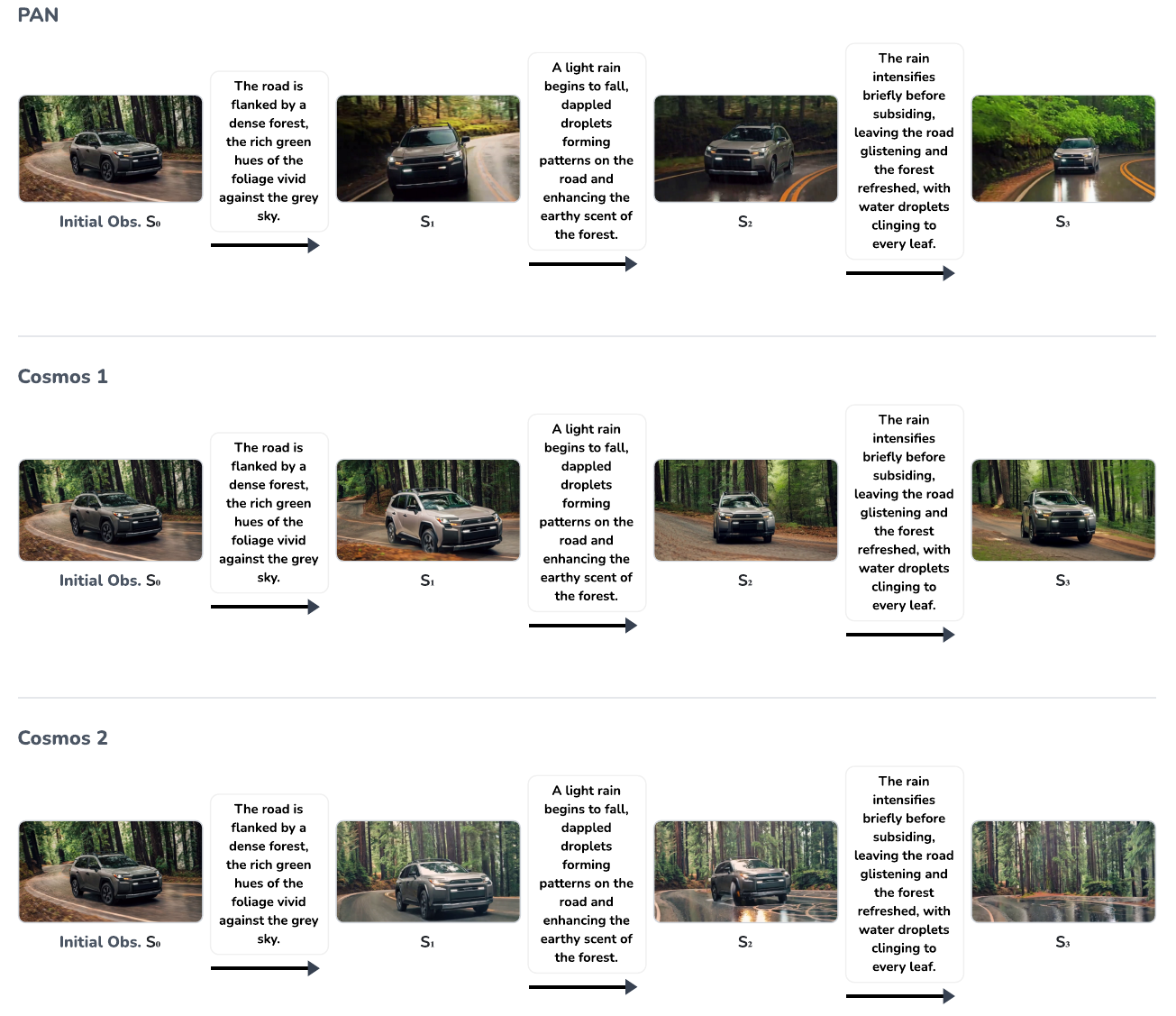}
    \caption{\textbf{Qualitative comparison on Simulation Fidelity.} Given a shared initial driving scene ($S_0$), PAN, Cosmos 1, and Cosmos 2 simulate a three-step sequence of environmental interventions from a dry forest road to light rain to intensifying rainfall.}
     \label{fig:case-action}
\end{figure}
\subsection{Simulation Fidelity}
Figure~\ref{fig:case-action} illustrates the \emph{environment simulation fidelity} dimension using a driving scenario with controlled interventions: starting from the same dry forest-road scene ($S_0$), the models simulate a three-step transition from dry conditions to light rain and then to intensifying rainfall. This setting probes whether a world model can preserve scene identity (vehicle, viewpoint, road geometry) while applying physically grounded, temporally consistent changes driven by the intervention description.

PAN demonstrates particularly strong intervention-following fidelity and scene continuity. Across steps, it maintains a stable camera viewpoint and consistent vehicle/road structure while introducing increasingly salient weather cues that align cleanly with the textual interventions. The transition from dry conditions to light rain is rendered in a controlled, believable manner, and the subsequent intensification produces a clear escalation in moisture effects—most notably through enhanced road specularity, richer glistening highlights, and an overall refreshed forest appearance consistent with rainfall accumulation. Importantly, PAN’s rollouts preserve the semantic identity of the scene while expressing the intervention through physically meaningful visual signals, yielding a smooth and realistic temporal evolution that supports reliable counterfactual evaluation.
In contrast, Cosmos~1 and Cosmos~2 exhibit weaker fidelity under the same intervention sequence. Their simulated transitions are less temporally structured, and the rendered changes can be less tightly coupled to the intended progression in rainfall intensity. In several steps, Cosmos rollouts show greater variability in appearance and scene rendering that is not clearly attributable to the intervention itself, which makes the causal effect of light rain versus intensifying rainfall harder to isolate. Additionally, compared with PAN, Cosmos outputs may under-express key physical cues (\emph{e.g.,} consistent wet-road reflectance and coherent atmospheric rain effects) or exhibit shifts in rendering that reduce continuity across time steps.

Overall, this case study highlights that simulation fidelity requires simultaneously satisfying two constraints: (i) \emph{identity preservation} (keeping geometry, viewpoint, and key entities consistent), and (ii) \emph{faithful intervention realization} (introducing the correct magnitude and type of physical change). Models that achieve both are better suited for realistic long-horizon rollouts and robust evaluation of environment-dependent decision making.

\section{Conclusion}
We have introduced \benchmark, a comprehensive benchmark designed to evaluate world models (WMs) as next world simulators — internal hypothetical engines that support reasoning, long-range forecasting, and purposeful action. Unlike prior evaluations that emphasize short-term prediction or perceptual fidelity, \benchmark systematically probes three advanced simulation capabilities: Action Simulation Fidelity, Long-horizon Forecast, and Simulative Reasoning and Planning. By curating diverse tasks and datasets that require models to follow high-level control, sustain coherent multi-step rollouts, and compare alternative futures, our benchmark shifts evaluation toward the functional roles that WMs must fulfill in real-world intelligence.

Through large-scale experiments, we find that current state-of-the-art WMs exhibit significant gaps in faithfully following high-level instructions and maintaining long-horizon simulation consistency, with error accumulation degrading multi-turn rollouts and environment-centric control proving particularly difficult. Only models that produce semantically actionable rollouts and jointly optimize understanding, prediction, and control, meaningfully improve planning performance.

By revealing these limitations, \benchmark serves not only as a diagnostic tool but also as a roadmap for future research. We hope that it will guide the development of next-generation world models capable of truly understanding, forecasting, and planning across diverse, real-world environments.

\clearpage
 
\appendix

\section{Contributors}

Qiyue Gao*, Kun Zhou*, Jiannan Xiang*, Zihan Liu*, Dequan Yang, Junrong Chen, Arif Ahmad, Cong Zeng, Ganesh Bannur, Xinqi Huang, Zheqi Liu, Yi Gu, Yichi Yang, Guangyi Liu, Zhiting Hu, Zhengzhong Liu, Eric Xing

\def\thefootnote{*}\footnotetext{Equal contribution.}\def\thefootnote{\arabic{footnote}}








\newpage
\bibliographystyle{plainnat}
\bibliography{references}


\end{document}